\documentclass[a4paper, 12pt]{article}
\usepackage[utf8]{inputenc}
\usepackage[margin=1in]{geometry}
\usepackage{array}
\usepackage{tabularx}
\usepackage{listings}
\usepackage[dvipsnames]{xcolor}
\usepackage{amsmath,amssymb}
\usepackage{bbm}
\usepackage{graphicx}
\usepackage{forest}
\graphicspath{ {./} }
\usepackage{setspace}
\usepackage[T1]{fontenc}
\usepackage{multirow}
\usepackage{array,calc,tabularx}
\usepackage{lastpage}
\usepackage[owncaptions,tablegrid]{vhistory}
\usepackage{lipsum}
\usepackage{float}
\usepackage{algorithm}
\usepackage{algpseudocode}
\usepackage{subcaption}

\newcommand{\HRule}{\rule{\linewidth}{0.5mm}}

\DeclareMathOperator*{\argmax}{arg\,max}

\begin{document}

\begin{center}

\begin{spacing}{0.3}
\noindent
\HRule
\end{spacing}

\vspace{1cm}

{\huge \textbf{Analyzing Hong Kong's Legal Judgments from a Computational Linguistics point-of-view}}

\vspace{0.4cm}

\begin{spacing}{0.0}
\noindent
\HRule\\
\HRule \\
\end{spacing}

\vspace{1cm}

Sankalok Sen$^{\clubsuit \diamondsuit}$\footnote[1]{Work done as a part of the author's Research Assistantship at The University of Hong Kong from July 2022 - May 2023.}

$^{\clubsuit}$\textit{Department of Computer Science, The University of Hong Kong}

$^{\diamondsuit}$\textit{ssen2001@connect.hku.hk}

\end{center}

\begin{abstract}

Analysis and extraction of useful information from legal judgments using computational linguistics was one of the earliest problems posed in the domain of information retrieval. Presently, several commercial vendors exist who automate such tasks. However, a crucial bottleneck arises in the form of exorbitant pricing and lack of resources available in analysis of judgements mete out by Hong Kong's Legal System. This paper attempts to bridge this gap by providing several statistical, machine learning, deep learning and zero-shot learning based methods to effectively analyse legal judgments from Hong Kong's Court System. The methods proposed consists of: (1) Citation Network Graph Generation, (2) PageRank Algorithm, (3) Keyword Analysis and Summarization, (4) Sentiment Polarity, and (5) Paragrah Classification, in order to be able to extract key insights from individual as well a group of judgments together. This would make the overall analysis of judgments in Hong Kong less tedious and more automated in order to extract insights quickly using fast inferencing. We also provide an analysis of our results by benchmarking our results using Large Language Models making robust use of the HuggingFace ecosystem.


\end{abstract}

\section{Introduction}

 In the following sections, we provide a brief history of Hong Kong's Legal System, the importance of this paper from an academic standpoint, engineering choices considered for this paper, and finally the overall objectives of what this paper attempts to accomplish. 

 \subsection{Brief History of Hong Kong Legal System}

 The Judicial Branch of the Hong Kong Special Administrative Region of the People’s Republic of China (HKSAR) exercises its control over the judicial needs and requirements of the region. It is independent from the influences of the Legislative and Executive Branches as conformed by the mandates of the Basic Law [1]. The Basic Law was ratified by the National People’s Congress on April 4, 1990, coming into effect after the handover of the region by 
the United Kingdom, on July 1, 1997. It replaced the Colonial Rules consisting of the Hong Kong Letters Patent and Hong Kong Royal Instructions of 1917 [2]. Broadly, the Courts of Law in Hong Kong which rule Judgments under the protection of the Basic Law, broadly consist of 8 different types as stated in Table 1.

\begin{table}[H]
\begin{center}
\begin{tabular}{|m{10cm}|} 
 \hline
 \textbf{Courts of Law}\\  
 \hline\hline
 1. Court of Final Appeal\\ 
 \hline
 2. Court of Appeal of the High Court\\
 \hline
 3. Competition Tribunal\\
 \hline
 4. District Court\\
 \hline
 5. Family Court\\
 \hline
 6. Lands Tribunal\\
 \hline
 7. Others: Magistrates' Court, Labour Tribunal, Small Claims Tribunal, Obscene Articles Tribunal, Coroner's Court\\
 \hline
\end{tabular}
\end{center}
\caption{\label{tab:table-nameC} Categories of Courts of Law in Hong Kong}
\end{table}

\subsection{Background \& Motivation}

Each of the Courts of Law as mentioned in Table 1 metes out multiple judgments every week. To facilitate legal research and impart academic teaching, law faculties in Hong Kong 
need to constantly update their database with respect to how each of these judgments differ from each other in case type, important, and wording. It is a manually exhaustive process, and several commercial third-party companies provide computationally automated solutions [3]. However, it is monetarily expensive and often such corporate solutions are not provided for Hong Kong’s judgments.

Therefore, this paper suggests a solution by combining several computational techniques in Natural Language Processing. This makes it easier to effectively analyse newer legal judgments from both unsupervised and supervised learning based points of view. The methods implemented consists of: (1) Citation Network Graph based Knowledge Generation, (2) PageRank Algorithm, (3) Keyword Analysis and Summarization, (4) Sentiment Polarity, and (5) Paragrah Classification, and to be able to extract key insights from individual as well a group of judgments together. 

\subsection{Engineering Choices}

In recent years, with release of more powerful computing processors, various papers were published which leveraged the theory of Deep Learning Models. The intuition behind Deep Learning comes from the structure of the human brain composed of neurons. Just like the human brain learns from new experiences, a deep learning model is said to be learning and mimicking a human neuron. Specific to NLP, huge advancements were seen with the proposal of the Attention Mechanism in 2014 by Bahdanau et al [4] and subsequent introduction of the Transformers Architectures by Vaswani et al in 2017 [5], both leveraging Deep Learning 
Models. These models witnessed a surge in improvements in natural language understanding tasks like Text Summarization, Generation, Sentiment, and Question-Answering Tasks, among others.

However, when these models are applied to a specific social science based domain to understand it better, they tend to generalize as they often constitute very large pre-trained models which are often trained on extremely generic datasets which do not fit well with the nature of the social science domain. Thus, this paper has chosen to adopt more general Probabilistic Machine Learning Models instead. All the three models as proposed in this paper, are based on this theory. In comparison, Deep Learning is said to be a more niche subset of Machine Learning, where the model seems to mimic a human neuron. This approach of choosing more general Machine Learning Models over specific Deep Learning Models is often taken when attempting to solve social science based problems as seen in [6], [7], [8], [9], [10], [11].

Therefore, this paper evaluates by comparing the proposed methods with the results of Deep Learning Models (specifically, Large Pre-trained Language Models), and using this analysis concludes which model is suitable for which parts of a task from a precision and accuracy perspective.

\subsection{Objectives}

Keeping these in mind, the paper presents its objectives:

\begin{itemize}
    \item Implementation of the Citation Network Model based Knowledge Graph for each judgment which have citations, to find more important citations in the past 25 years and causal connections between various judgments since the handover of Hong Kong. 
    \item Implementation of the Google’s PageRank Algorithm [12] and apply it to the already built Citation Network Models to find the top citations among the data available to provide an overall score to each judgment.
    \item Implementation of Keyword Analysis Algorithms ({\sc TextRank} [13], {\sc YAKE} [14], {\sc RAKE} [15], {\sc KeyATM} [16], {\sc LDA} [17]), to extract keywords and phrases as well as effectively summarise each judgment and benchmarking using BERT (base) for benchmarking.
    \item Implementation of Sentiment Analysis ({\sc VADER} [16]) to extract sentiment distribution across a judgment paragraph-wise.
    \item Implementation of a Paragraph level classifier for each Judgment improving the quality of semantic extraction for each paragraph using algorithms like Naive Bayes, Support Vector Machines (SVMs), Bernoulli Restricted Boltzmann Machines (BernoulliRBM), Stochastic Gradient Descent (SGD), Multilayer Perceptrons (MLP), and BART trained on Multi-Genre Natural Language Inference for benchmarking, using One to Few-Shot Learning. 
\end{itemize}

\section{Literature Review}

In the following section we go through an extensive review of work done in the field of Legal Judgment Analysis using Machine Intelligence in the past two decades. 

[19] introduced a parallelly translated corpus of Italian and German legal texts which are used for testing different translation methods like alignment architectures. [20] used a corpus compiled from House of Lords (United Kingdom) judgments which are used for creating effective summarisation techniques in relation to legal judgments. [21] experimented with a question-answering method for creating a human-like tool used for testing against the Bar examination held in the United States of America, stating the Bar preparation materials as the legal corpus. [22] examined various methods posed by competitors in a case law information extraction competition on Japanese legal documents. [23] introduced a dataset which are used for testing Named Entity Recognition tasks on Brazillian legal texts. [24] introduced a dataset consisting of more than 2.8 million criminal cases as released by the Supreme Court of China in attempts to test for judgment prediction using both Machine Learning and Deep Learning techniques and providing benchmark comparisons between the two techniques. [25] provided a dataset of different legal contracts and an attempt to summarise them as informal English for the ease of human language understanding. [26] introduced a dataset compiled from judgments by the United Nations Convention on Humar Rights and attempted to predict results using different neural techniques. [27] introduced a large dataset of 57K judgments in European Union and attempts a robust multi-label classification method. [28] introduced a dataset of 10K judgments on Chinese judicial reading along with 50K questions and answers and tests methods like {\sc BERT} and {\sc TF-iDF} compared to human annotated benchmarks. 

[29] randomly selected 50K Chinese Judgments as published online by the Courts of China and implements single and multi-level classification as well as Bi-Directional GRU architectures and models a charge prediction task for criminal cases. [30] also tackles the charge prediction task but also used Positional Embeddings, Part of Speech Tags, Bigram Models and WordNet on top of their deep neural architectures and received better prediction accuracies. [31] used an Encoder with Attention towards prediction of accurate judgments for legal reading comprehension texts. [32] implemented Markov Network Models to predict judgments in relation to divorce cases.  

[33] attempted to analyse criminal cases for tackling the courts view generations task using a label-conditional sequence-to-sequence model with attention. [34] tackles the same problem using an attention based encoder architecture and an innovation counterfactual decoder architecture with pointer-generator. [35], [36], [37] attempted to extract legal entities using different Named Entity Recognition techniques. [38], [39] attempted to extract events from legal texts using techniques like temporal reasoning. [40] tried information retrieval techniques on legal judgments using paragraph and citation information. [41] built a pre-trained phrase scoring model for information retrieval using summarization and lexical matching techniques. [42] used a combination of rule-based and statistical methods to first create an automatic summarization tool for legal judgments. [43] used the LDA algorithm in an attempt to summarize legal documents. [44] leveraged domain knowledge methods towards legal text summarization effectively. 

\section{Programming Languages \& Tools}

This project leverages modern powerful processors by utilizing resources of the GPU Farm of the Department of Computer Science, The University of Hong Kong to attain its results. The main programming languages used for coding the models and algorithms is Python.

The justification for the choice of Python is that it can easily handle big data and able to easily fit Language models on the dataset.
The scripting is done in Jupyter Notebooks and executed on the GPU Farm. The choice of using Jupyter Notebooks is because it provides an easy interface for interactive simulation. For instance, for a given dataset, it is very easy visualize different types of outputs in form of graphs, tables, and charts. GPU (Graphics Processing Unit) is a special type of circuit which speeds up numerical computation, and so the GPU Farm of HKU has been used to speed up the overall computation of the models.

\section{Dataset}

The {\sc Legal-NLP} Dataset was extracted as a part of another ongoing project at the Natural Language Processing Lab, Department of Computer Science, HKU (HKUNLP), as is situated in the departmental server of the project supervisor, with each Judgment extracted and stored 
as a JSON File, with each JSON File structured as shown in Table 2. 

\begin{table}[H]
\begin{center}
\begin{tabular}{|m{3cm}|m{11cm}|} 
 \hline
 \textbf{Key Value} & \textbf{Pair Values} \\  
 \hline\hline
 judgement & $[data_i, [type = \{other, para, heading, quote\}]]$ $\forall i$ (List of Lists) \\ 
 \hline
 ref & $[ref_i]$ $\forall i$ (List of References)  \\
 \hline
 date & $[date_i]$ $\forall i$ (All Hearing Dates)\\
 \hline
 parties & $[party_i]$ $\forall i$ (Parties in Dispute)\\
 \hline
 coram & $[coram_i]$ $\forall i$ (\textit{Coram Judice})\\
 \hline
 representation & $[rep_i]$ $\forall i$ (Representations for all parties)\\
 \hline
 caseno & [caseno] (Extracted Case-No)\\
 \hline
\end{tabular}
\end{center}
\caption{\label{tab:table-nameR} Document Structure of each JSON holding one Judgment}
\end{table}

Before being preprocessed into a JSON File, it was downloaded from the Hong Kong Legal Information Institute (HKLII) [45]. Preprocessing was done using semantic parsing and regular expressions. The {\sc Legal-NLP} Dataset is formally stated in this subsection. Mathematically, the Dataset $\mathcal{D}$ consists of Judgements $|\mathcal{J}|$, which can be described as:

\begin{align*}
    \forall c = \{c_1, c_2, ... , c_M\} &\in \mathcal{C} \text{  (Courts of Law)}\\
    \exists j = \{j_{c,1}, j_{c,2}, ..., j_{c,N}\} &\in |\mathcal{J}|_{C,N} \text{ (\#\{Judgments per Court\})}\\
    \text{such that  } \sum_{c=1}^{c=M}\sum_{j=1}^{j=N} J_{c_m, j_n} &= |\mathcal{J}| =  \mathcal{D} \tag{Equation 1}
\end{align*}

The Dataset consists of approximately 115,000 Bilingual Judgments, with around 80,000 in English and remaining 35,000 in Traditional Chinese. The Categories of Courts of Law as described in Table 1 can be divided in 28 semi-broad entities, with cases existing for about 18 semi-broad entities in the Dataset. A graph summarizing this distribution has been shown in Figure 1.

73 different types of cases were identified out of a total possible 112 [46], each of which can be described using its own unique code for identification. The total number of words in the Dataset is approximately 251 million with an average of 3000 words per judgment.

\begin{figure}[H]
\centerline{\includegraphics[width=14cm]{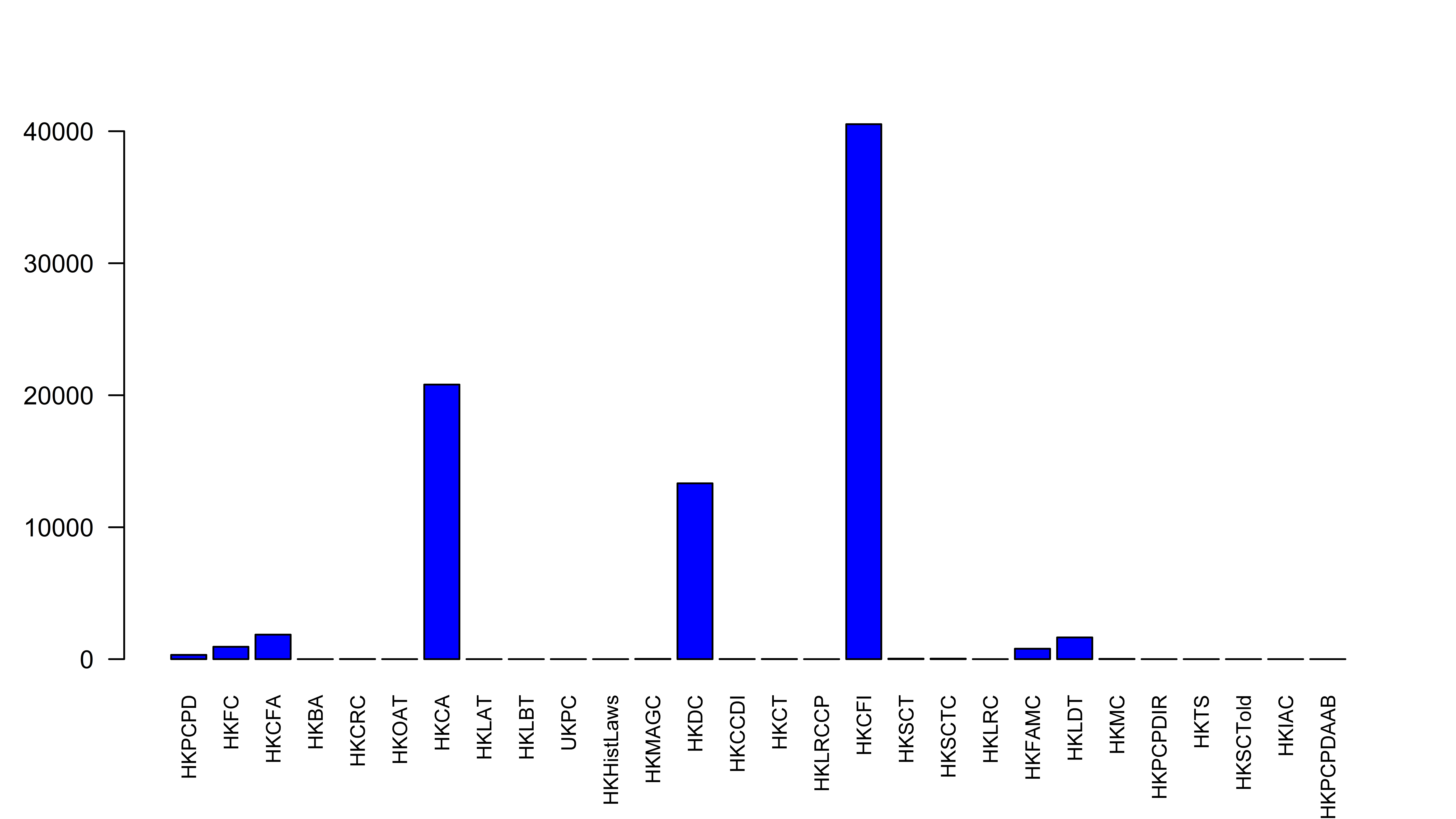}}
\caption{Case Counts vs. Hong Kong Courts}
\end{figure}

\section{Methodology}

\subsection{Citation Network based Knowledge Graph and PageRank}

Each Judgment has a specific number of citations, with pre-handover cases being rejected to concentrate the date over the specific 25 years period. The citations were extracted using various syntactical methods based off their citation styles including both brute force approach and regular expressions, resulting in a key-value pair based dictionary data-type. A Key-Value Dictionary is such that for every unique Key (Judgment Citation Number), there exist a certain number of citations existing in a list like format which are described as Values.

After cleaning the prepared Graph, to make it more structured owing to the different citation styles, the PageRank Algorithm was implemented. The PageRank Algorithm is a re-implementation of the "Google Algorithm" [12] - the official algorithm that the Google search engine uses to rank its web pages based on certain searches. This was implemented on the Graph Citation Network to view the most important cases in Hong Kong’s history post handover having the most present day impact.

Then, we color the Citation Network Graph using the following rules. The various colors are described as: Red [Lead Major Case being viewed], Blue [Citing Red], Green [Citing Blue], Yellow [Citing Green], Purple [Citing Yellow], Pink [Citing mixture of different cases at different levels]. Pink cases is significant cause it helps 
us track ’transitioning’ cases, i.e., cases that not only have cited the lead parent cases but also some other child case of the parent case signifying that there have been ideological shifts in meting out a particular judgment to such similar cases over time resulting in a judgment not only taking inspiration from the parent case but also its different child cases that have cited the parent case.

Algorithm 1 as shown below displays the creation of the {\sc State-Space-Graph}.

\begin{algorithm}[H]
\caption{{\sc State-Space-Graph} Generation}\label{alg:cap1}
\begin{algorithmic}
\footnotesize
\Require List of Cases each of DICT() type $\mathbb{C} = \{c_1, c_2, ..., c_N
\}$
\Ensure State Space Graph $\mathbb{G} = $  DICT()
\For{$c_i$ in $\mathbb{C}$}

\If{$c_i$[CASE NUMBER] {\bf not in} $\mathbb{G}$.KEYS()}
    \State $\mathbb{G}$[$c_i$[CASE NUMBER]] = []
\EndIf

\For{$r_i$ {\bf in} $c_i$[LIST OF REFERENCES]}
\If{$r_i$ {\bf not in} $\mathbb{G}$.KEYS()}
    \State $\mathbb{G}$[$r_i$] = [$c_i$[CASE NUMBER]]
\Else
    \State $\mathbb{G}$[$r_i$].APPEND($c_i$[CASE NUMBER])
\EndIf
\EndFor
\EndFor
\Require $\mathbb{K}$ = $\mathbb{G}$.KEYS()
\For{$k$ {\bf in} $\mathbb{K}$}
\State $\mathbb{G}$[$k$] = LIST(SET($\mathbb{G}$[$k$]))
\EndFor
\end{algorithmic}
\end{algorithm}

This {\sc State-Space-Graph} consists of key-value pairs using which we generate an Acyclic Directed Graph consisting of 12068 nodes and 12663 edges. We report a graph density of 8.695648829995403e-05. We apply the PageRank Algorithm using the default value of $0.85$ for the damping factor, which is the damping parameter.

Similar to webpages, each judgment case is denoted as a graph in the web of the internet (or, in this case the Hong Kong Legal System). As PageRank itself was inspired by Academic Citation Analysis as stated in the eponymous paper, we concluded that implementing it for Judgment Citation Analysis would provide us with good results due to the similar nature of the task to be tackled. 

After this, we implement Algorithm 2, which helps us to generate a colored citation network graph which is described as stated in {\sc Citation-Network-Coloring} given a particular CASE-NUMBER.

Using this graph, as derived using Algorithm 2, we correlate the positions 0 with RED, 1 with BLUE, 2 with GREEN, 3 with YELLOW and 4 with PURPLE color for each node. If a node has multiple positions attached to it that means it has been cross-cited, and so it is given the color PINK. 

\begin{algorithm}[H]
\caption{{\sc Citation-Network-Coloring}}\label{alg:cap2}
\begin{algorithmic}
\footnotesize
\Require CASE-NUMBER, sub-graph-edges = [], tree-structure = DICT(), tree-structure[0] = [CASE-NUMBER], tree-structure[1] = [], tree-structure[2] = [], tree-structure[3] = [], State Space Graph $\mathbb{G}$

\Ensure $\mathbb{L}$ = LIST(SET(LIST($\mathbb{G}$.PREDECESSORS(CASE-NUMBER))))

\For{$l$ {\bf in} $\mathbb{L}$}
\State sub-graph-edges.APPEND(($l$, CASE-NUMBER))
\If{$l$ {\bf not in} tree-structure.KEYS()}
\State tree-structure[$l$] = [$l$]
\Else
\State tree-structure[$l$].APPEND($l$)
\EndIf
\EndFor

\Ensure depth = 2, loops = 1

\While{$\mathbb{L}$ != [] {\bf or} depth > 3}
\For{$l$ {\bf in} $\mathbb{L}$}
\State predecessors-l = LIST(SET(LIST($\mathbb{G}$.PREDECESSORS($l$))))
\State $\mathbb{L}_{\text{update}}$ = []
\For{$i$ in predecessors-l}
\State sub-graph-edges.APPEND(($i$, $l$))
\If{depth {\bf not in} tree-structure.KEYS()}
\State tree-structure[depth] = [$i$]
\Else
\State tree-structure[depth].APPEND($i$)
\EndIf
\EndFor
\EndFor
\State $\mathbb{L}_{\text{update}}$.APPEND($i$)
\State depth = depth + 1
\State loops = loops + 1
\EndWhile

\Ensure New Graph $\mathbb{G}_{\text{new}}$
\State $\mathbb{G}_{\text{new}}$.ADD-EDGES-FROM(sub-graph-edges)

\end{algorithmic}
\end{algorithm}

\subsection{Keyword Analysis and Summarization}

We perform the following algorithms for visualization for some of the top cases extracted from the PageRank, and then benchmark between each other to perform the similarity of their results for evaluation on the HKCFA (Hong Kong Court of Final Appeals) subset of judgments. 

\subsubsection{TextRank}

The TextRank paper [13] proposed an innovative method for keyword extraction and text summarization by converting a text into a graphical structure, and hence choosing key linguistic structures by methods of voting and recommendation similar to that shown in PageRank using the same scoring index.

Given a document $\mathcal{D}$, perform tokenization, and then construct a graph based on the tokenized text. Then, rank the graph using the PageRank scoring mechanism, where for a vertex $V_j$ in the constructed graph, $In(V_j)$ is the set of vertices that point to the predecessor vertices, and $Out(V_j)$ is the set of vertices that point
to to the successor vertices. The score for a vertex is defined as:

\begin{align*}
    S(V_i) &= (1-d) + d\sum_{j \in In(V_i)} \frac{1}{|Out(V_j)|} S(V_j)
\end{align*}

The parameter $d$ is the damping factor, and is default parameterised as $0.85$ as suggested by the eponymous PageRank paper.

\subsubsection{Rapid Automatic Keyword Extraction (RAKE)}

The RAKE paper [15] proposed a method of keyword extraction by partitioning the document using punctuation and stop-words, and construct word-level co-occurrence matrices and use the
computed word scores to extract the top words.

The initial candidate keywords are selected as phrases occurring in-between stop-words or phrase delimiters, and then followed by graph construction using co-occurrences of keywords. The degree of a word $w$ is $deg(w)$ is computed from the constructed graph and the word frequency is computed as $freq(w)$, finally computing the ratio as $\frac{deg(w)}{freq(w)}$. For each candidate phrase initially selected, the scoring is computed by summing up the individual scores for each keyword phrase:

\begin{align*}
    S(\text{cand-phr}) &= \sum_{w \in \text{cand-phr}} S(w)
\end{align*}

Finally, it takes into multi-occurring stopwords to include in the candidate keyword phrase, and produces a final scoring and ranking them in descending order of scoring. 

\subsubsection{Yet Another Keyword Extractor (YAKE)}

The YAKE paper [14] is another recent keyword extractor inspired by RAKE which used additional statistical estimators as features of extraction consisting of Position of Words, Word frequency, Term Relatedness to Context, Term Different Sentence structures. This was
followed by term scoring, deduplication and final re-ranking.

The scoring of candidate keywords is given as follows:

\begin{align*}
    S(kw) &= \frac{\prod_{t \in kw} S(t)}{KF(kw) \times (1 + \sum_{t\in kw} S(t))}
\end{align*}

The $kw$ represents a 1 or more $n$-gram keyword for scoring, $S(t)$ are the candidate $n$-gram probability scores constituting the candidate keyword, and $KF(kw)$ is the candidate keyword's overall keyword frequency as described in the paper. 

\subsubsection{Latent Dirichlect Allocation (LDA)}

The LDA paper [18] describes generating topics from a corpus of texts. The mathematical description based on the paper is described as the following. 

For a given corpus of $N$ documents, each with length $n_i$, select $\theta_i \sim Dir(\alpha)$, with a sparse parameter $\alpha < 1$ for $i = \{1,...,N\}$. To extract $K$ topics, select $\phi_k \sim Dir(\beta)$ for a sparse parameter $\beta < 1$ for $k = \{1,...,K\}$. For each word position in the document and length of document $i \in \{1, ..., M\}$, $j \in \{1, ..., N_i\}$ respectively, compute firstly, the topic $t_{i,j} \sim Multinomial(\theta_i)$ and the word $w_{i,j} \sim Multinomial(\phi_{t_{i,j}})$.

\subsection{Sentiment Analysis}

We use the Valence Aware Dictionary for Sentiment Reasoning (VADER) [16] paper to extract the linguistic sentiment of each paragraph in a judgment for gaining insights into the sentiment variate of the judge while meting out a judgment. Based on human checking, most cases describe the admissal or dismissal of the plaintiff or defendant's case in the beginning or right at the ending paragraphs for a judgment. Finally, we visualize this by plotting the paragraph-wise sentiment variate distribution in a graph.

\textit{\underline{Limitations:} This approach highlights the lack of distinguishing sympathetic or other raw emotional variations in documents. This can be solved by adding a tagger to the paragraph by means of text classification, to decipher paragraphs from each other. For example, a paragraph might go as "Let us sympathize with the victim's family...", might be tagged a positive linguistic sentiment. However, in the context of the overall case, there might be an ambiguity in analysis of results. However, if the tagger descibes it to be a part of Description/About the case instead of a judge's Opinion/Ruling, this would greatly improve the understanding of the corpus text for a legal researcher/academic. We propose multiple methods for paragraph-wise text classification in the following subsection to tackle this challenge posed and propose a benchmarking method for analyzing our results, and finally suggest an innovative Zero-Shot Learning based approach for quick inferencing. This makes use of the recent surge of large language models which are often trained on a large volume of data and helpful in generalizing results better than trained models due to its better understanding of linguistic substructures in textual data.} 

\subsection{Paragraph-wise Text Classification}

In the following section, we implement various Machine Learning and Deep Learning Algorithms for paragraph-wise Judgment Text Classification task. After careful inspection of individual judgments, we classified a subset of 50 documents and extracted their paragraphs into 4 types:

\begin{itemize}
    \item \textbf{About:} Description of the case by the Judge.
    \item \textbf{Ruling:} Describes a neutralized ruling or opinion of the Judge.
    \item \textbf{Allowed:} Describes an opinion of the judge in favour of the plaintiff or defendant.
    \item \textbf{Dismissal:} Describes an opinion of the judge against the plaintiff or defendant.
\end{itemize}

\subsubsection{Naive Bayes}

\begin{figure}[H]
\centerline{\includegraphics[]{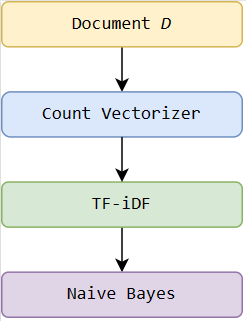}}
\caption{Naive Bayes Model}
\end{figure}

The above figure shows the Naive Bayes Model implemented. The count vectorizer aids in conversion of a text document corpus to a matrix of token counts. The Term Frequency–inverse Document Frequency or TF-iDF for a word/term $t$, individual document $d$, cluster of documents $D$, total number of documents $|D|=N$, and number of documents for which a term $t$ appears as $|d\in D:t\in d|$,  is defined as:

\begin{align*}
    TF(t,d) &= \frac{freq(t,d)}{\sum_{t'\in d} freq(t',d)} \\
    iDF(t,D) &= \log \frac{N}{1+|d\in D:t\in d|} \\
    TF\text{-}iDF(t,d,D) &= TF(t,d) \cdot iDF(t,D)
\end{align*}

To perform Naiver Bayes classification, we wish to predict a class $\hat{y}$ for a sentence $(x_1, x_2,...,x_n)$:

\begin{align*}
    \mathbb{P}(y|x_1,x_2,...,x_n) &\propto \mathbb{P}(y)\prod_{i=1}^{n} \mathbb{P}(x_i|y) \\
    \hat{y} &= \argmax_i \mathbb{P}(y)\prod_{i=1}^{n} \mathbb{P}(x_i|y)
\end{align*}

For the priors, we consider three types of Naive Bayes submodels: (1) Bernoulli, (2) Multinomial, and (3) Complement (Multinomial for sparse datasets).

\subsubsection{Linear Support Vector Machine}

\begin{figure}[H]
\centerline{\includegraphics[]{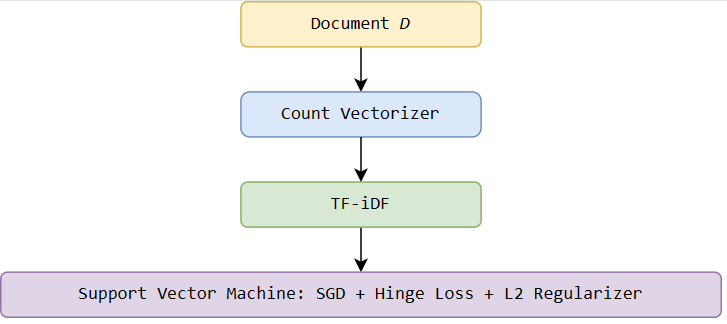}}
\caption{Support Vector Machine}
\end{figure}

The above figure shows the Support Vector Machine implemented. For a given set of sentences and classes $(\vec{X}, Y)$, we wish to linearly separate the data based on respective classes. To perform this tast, we use Stochastic Gradient Descent, with Hinge Loss and $L_2$ regularizer. A hyperplane separating the data into clusters can be defined by $w^T \cdot \vec{X} - b = 0$. With a parameter $\lambda > 0$, the optimization problem is theoretized as:

\begin{align*}
    \min \left(\lambda||w||_2^2 + \frac{1}{n}\sum_{i=1}^n \max(0, 1-y_i(w^T \cdot x_i - b))\right)
\end{align*}

Solving this classifies the data into required classes, and helps to evaluate our dataset on the fitted SVM model.

\subsubsection{Logistic Regression}

\begin{figure}[H]
\centerline{\includegraphics[]{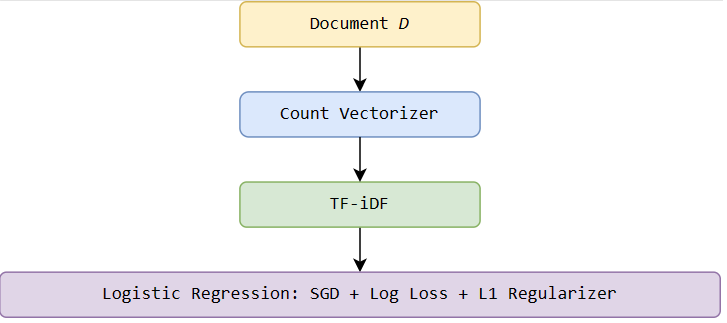}}
\caption{Logistic Regression}
\end{figure}

The above figure shows the Logistic Regression model implemented. We compute the priors as:

\begin{align*}
    p_X(\vec{x_i}) = \mathbb{P}(\vec{x_i}) &= \frac{1}{1+ e^{-\boldsymbol{\beta} \vec{x_i}}}
\end{align*}

We wish to minimize the log likelihood estimate and predict the classes as:

\begin{align*}
    C &= \min \sum_{i=1}^{n} y_i \log (p_X(\vec{x_i})) + (1 - y_i) \log (1-p_X(\vec{x_i}))
\end{align*}

\subsubsection{Beroulli Restricted Boltzmann Machine}

\begin{figure}[H]
\centerline{\includegraphics[]{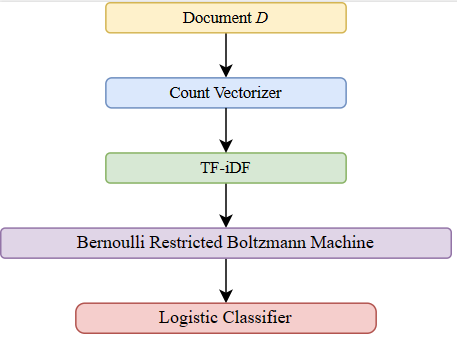}}
\caption{Restricted Boltzmann Machine}
\end{figure}

The above figure shows the Restricted Boltzmann Machine implemented with a Logistic Classifier.

It consists of visible $v$ (with offsets $a$) and hidden $h$ (with offsets $b$) binary units and is a stochastic generative neural network and a weight matrix $w_{i,j} \in W$. The classification model is computed as:

\begin{align*}
    E(v,h) &= -a^Tv - b^Th - v^TWh \\
    \mathbb{P}(v,h) &= \frac{e^{-E(v,h)}}{\sum e^{-E(v,h)}} \\
    \mathbb{P}(h_j = 1 | v) &= \sigma\left(b_j + \sum_i w_{i,j}v_i \right) \\
    \mathbb{P}(v_i = 1 | h) &= \sigma\left(a_i + \sum_j w_{i,j}h_j \right) 
\end{align*}

We wish to maximize the following for an input sentence to classify it:

\begin{align*}
    \argmax_W \prod_{x \in \vec{X}} \mathbb{P}(x)
\end{align*}

\subsubsection{Base Linear Model with Embeddings}

\begin{figure}[H]
\centerline{\includegraphics[]{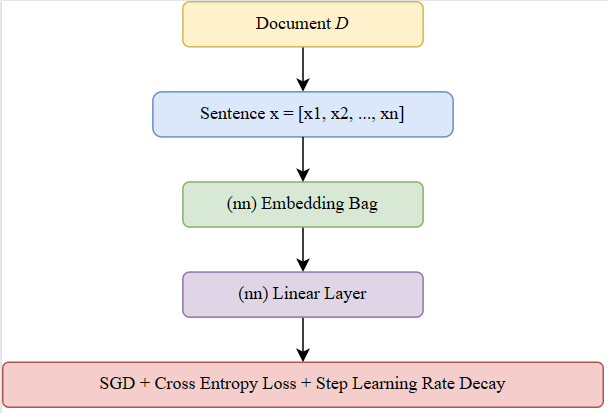}}
\caption{Linear Model with Embeddings}
\end{figure}

For the base deep learning model we create a simple model with Embeddings and Linear Layer. We use Stochastic Gradient Descent Algorithm with Cross Entropy Loss and Step Learning Rate Decay for optimization. The model is visualized in Figure 6.

\subsubsection{Encoder (LSTM) + Decoder (LSTM + SelfAttention)}

\begin{figure}[H]
\centerline{\includegraphics[]{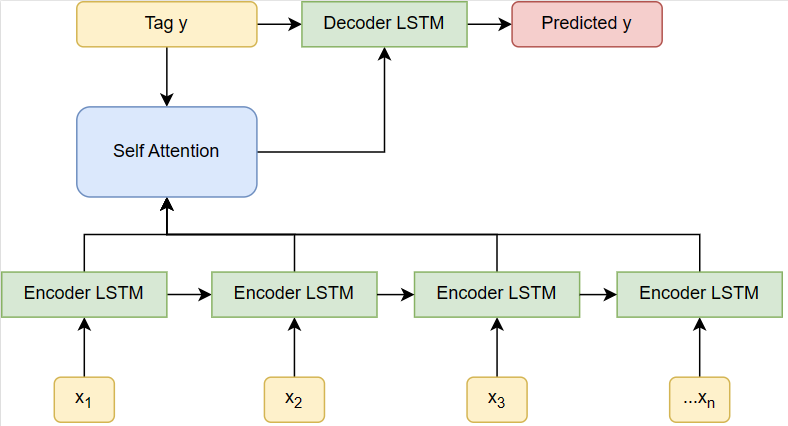}}
\caption{Encoder (LSTM) + Decoder (LSTM + SelfAttention)}
\end{figure}

For the improved deep learning model we create an LSTM based Encoder and an LSTM with Self Attention based Decoder. The model is visualized in Figure 7.

We compute Self Attention for Key ($K$), Values ($V$), and Query ($Q$) pairs as:

\begin{align*}
    \text{SelfAttention}(Q, K, V) &= \text{softmax}\left(\frac{QK^T}{\sqrt{dim(K)}}\right)V
\end{align*}

\subsubsection{BART Large MultiNLI for Paragraph Classification}

Finally, we compare our results with the pre-trained BART Large MultiNLI Large Language Model, created by Facebook AI Research, and show it's useful qualities for fast Paragraph-level Classification of Judgments.

\section{Results}

In the following subsections, we go through the results of our experiments and findings. Firstly, we state the results of the PageRank algorithm. Then, we show the results of the implemented Machine Learning, Deep Learning, and Zero-Shot Learning inference, and compare and contrast between different models. Finally, we present the results of Citation Network Graphs, Keyword Analysis, Summarization methodology for some of the top cases extracted from the PageRank Algorithm.

\subsection{PageRank Results}

The Top 20 cases extracted as stated in Table 3 below with their scoring from the PageRank Algorithm with default parameters.

\begin{table}[H]
\begin{center}
\begin{tabular}{|m{4.5cm}|m{4.5cm}|} 
 \hline
 \textbf{Case Number} & \textbf{PageRank Score}\\  
 \hline\hline
 7 HKCFAR 187 & 0.00676 \\
  \hline
CACV 284/2017 & 0.00232 \\
 \hline
CACV 54/2018 & 0.00221 \\
 \hline
CACV 219/2018 & 0.00199 \\
 \hline
2 HKLRD 1121 & 0.00162 \\
 \hline
1 HKLRD 69 & 0.00162 \\
 \hline
1 HKLRD 1 & 0.00136 \\
 \hline
3 HKLRD 691 & 0.00115 \\
 \hline
10 HKCFAR 676 & 0.00108 \\
 \hline
1 HKC 261 & 0.00106 \\ 
 \hline
2 HKLRD 437 & 0.00101 \\ 
 \hline
CACC 338/2007 & 0.00098 \\
 \hline
HCAL 106/2017 & 0.00096 \\
 \hline
5 HKLRD 1 & 0.00095 \\
 \hline
5 HKCFAR 356 & 0.00084 \\
 \hline
2 HKLRD 12 & 0.00071 \\
 \hline
2 HKLRD 1 & 0.00070 \\
 \hline
CACV 65/2014 & 0.00068 \\
 \hline
FACV No 16 of 2008 & 0.00066 \\

 \hline
\end{tabular}
\end{center}
\caption{\label{tab:table-namePR} PageRank (Top 20 Judgments)}
\end{table}

\subsection{Keyword Analysis Results}

In this section, we outline the results of our Keyword Extraction Models. We hypothesize that we can choose some model over the other if they have a close correlation and scoring with respect to other model or some model extracts a high quality and quantity of keywords with respect to other models. The 5 models implemented were TextRank, RAKE, YAKE, LDA (with a singular topic generated), and KeyBERT, with KeyBERT being the benchmark as it is a Large Language Model. We summarize our results in Table 4 and 5. 

\begin{table}[H]
\begin{center}
\begin{tabular}{|m{3.8cm}|m{2cm}|m{2cm}|m{2.1cm}|m{2cm}|m{1.9cm}|} 
\hline
\textbf{Metric/CN} & \textbf{HKCCDI}&\textbf{HKFAMC}&\textbf{HKMAGC}&\textbf{HKSCTC}&\textbf{HKOAT} \\  
\hline\hline
 \textbf{Case Count} & \textbf{8} & \textbf{749} & \textbf{20} & \textbf{37}& \textbf{1} \\
\hline
 TextRank-RAKE & 0.0902 & 0.1118&0.0924 & 0.1067& 0.1714 \\
\hline
 TextRank-YAKE & 0.1360 & 0.1900&0.1258 & 0.1448& 0.2553 \\
\hline
 TextRank-LDA & 0.0645& 0.1189& 0.0610& 0.0773& 0.1162 \\
\hline
 TextRank-KeyBERT  & 0.0160& 0.0279& {\color{red}0.0209} & 0.0216& {\color{red}0.0232} \\
\hline
 RAKE-YAKE & 0.1000& 0.0968&0.0787 &0.1120 & 0.1509 \\
\hline
 RAKE-LDA  &0.1250 & 0.0771 & 0.0582& 0.0944& 0.0869 \\
\hline
 RAKE-KeyBERT & {\color{red} 0.0000} & {\color{red}0.0160} & 0.0134 & {\color{red} 0.0186}& {\color{red}0.0000} \\
\hline
 YAKE-LDA & {\color{ForestGreen} 0.3043}& {\color{ForestGreen}0.3843} & {\color{ForestGreen}0.3699}& {\color{ForestGreen} 0.3516}& {\color{ForestGreen}0.2857} \\
\hline
 YAKE-KeyBERT & {\color{red} 0.0000} & 0.0575& 0.0840& 0.0670& 0.0454 \\
\hline
 LDA-KeyBERT  & {\color{red} 0.0000} & 0.0572& 0.0621& 0.0525& 0.0909 \\
\hline
\end{tabular}
\end{center}
\caption{\label{tab:table-nameKAM1} Keyword Analysis Metrics for each Court (Part 1)}
\end{table}

\begin{table}[H]
\begin{center}
\begin{tabular}{|m{3.8cm}|m{2cm}|m{2cm}|m{2cm}|m{2cm}|m{2cm}|} 
\hline
\textbf{Metric/CN} & \textbf{HKCFA}&\textbf{HKCRC}&\textbf{HKCT}&\textbf{HKFC}&\textbf{HKMC}\\  
\hline\hline
 \textbf{Case Count} & \textbf{1544} & \textbf{8} & \textbf{12} & \textbf{872}& \textbf{18} \\
\hline
 TextRank-RAKE & 0.1173 & 0.0902& 0.1026& 0.1127& 0.0966 \\
\hline
 TextRank-YAKE & 0.1899 & 0.1360& 0.1804& 0.1916& 0.1289 \\
\hline
 TextRank-LDA & 0.1452& 0.0645& 0.1252& 0.1215& 0.0627 \\
\hline
 TextRank-KeyBERT  & {\color{red} 0.0559}&{\color{red}0.0160}& 0.0302& {\color{red}0.0294}& {\color{red}0.0211} \\
\hline
 RAKE-YAKE & 0.2005& 0.1000& 0.1107& 0.1002& 0.0806 \\
\hline
 RAKE-LDA  &0.1448 &0.1250 & 0.1068& 0.0790& 0.0605 \\
\hline
 RAKE-KeyBERT & 0.0621 & {\color{red} 0.0000}& {\color{red}0.0207}& {\color{red}0.0165}& {\color{red}0.0148} \\
\hline
 YAKE-LDA & {\color{ForestGreen} 0.4348}& {\color{ForestGreen} 0.3043}& {\color{ForestGreen} 0.3983}& {\color{ForestGreen}0.3858}& {\color{ForestGreen}0.3610} \\
\hline
 YAKE-KeyBERT & 0.1280& {\color{red} 0.0000}& 0.0920& 0.0603& 0.0813 \\
\hline
 LDA-KeyBERT  & 0.1488 & {\color{red} 0.0000}&  0.0820& 0.0607& 0.0523 \\
\hline
\end{tabular}
\end{center}
\caption{\label{tab:table-nameKAM2} Keyword Analysis Metrics for each Court (Part 2)}
\end{table}

\textit{\underline{Conclusion:}} From both of the tables where we visualize the computed Keyword Analysis metrics, we conclude that \textbf{\color{teal} YAKE-LDA} have the most common percentage of outcome keywords between them. Whereas, \textbf{\color{red} KeyBERT} by itself have the least common percentage of outcomes, highlighting that KeyBERT extracts unique keywords different from traditional models.

Therefore, while considering different Keyword Analysis Algorithms, we must consider the tasks at hand and use a host of different algorithms and visualize the extracted keywords independently on average to extract unique insights into the Judgments.

\subsection{Summarization Results}

In the analysis of our summarization results, we use the Recall-Oriented Understudy for Gisting Evaluation (or ROUGE) [47] metric for our computation analysis. We visualize the Rouge-1, Rouge-2, and Rouge-L and state their Precision, Recall and F1-Scores. We benchmark the TextRank summarization against Facebook's BART-Large-CNN Model for the Summarization Task. 

With the statistics summarized in the following Table 6, we define them as shown below:

\begin{align*}
    \text{Precision} &= \frac{\text{True Positive}}{\text{True Positive} + \text{False Positive}} \\
    \text{Recall} &= \frac{\text{True Positive}}{\text{True Positive} + \text{False Negative}} \\
    \text{F1 Score} &= \frac{2\times\text{Precision}\times\text{Recall}}{\text{Precision} + \text{Recall}}
\end{align*}

\begin{table}[H]
\begin{center}
\begin{tabular}{|m{3.8cm}|m{2cm}|m{2cm}|m{2cm}|m{2cm}|m{2cm}|} 
\hline
\textbf{Metric/CN} & \textbf{HKCFA}&\textbf{HKCA}&\textbf{HKDC}&\textbf{HKFC}&\textbf{HKLDT}\\  
\hline\hline
 \textbf{Case Count} & \textbf{25} & \textbf{25} & \textbf{25} & \textbf{25}& \textbf{25} \\
\hline
 ROUGE-1 Recall & {\color{teal}0.3242} & {\color{teal}0.3236}& {\color{teal}0.2783}& {\color{teal}0.2057}& \textbf{\color{teal}0.2233} \\
\hline
 ROUGE-1 Precision & {\color{teal}0.4489} & {\color{teal}0.5060}& {\color{teal}0.3958}& {\color{teal}0.4545}& \textbf{\color{teal}0.6285} \\
\hline
 ROUGE-1 F1 & {\color{teal}0.3270}& {\color{teal}0.3541}& {\color{teal}0.3019}& {\color{teal}0.2762}& \textbf{\color{teal}0.3297} \\
\hline
 ROUGE-2 Recall  & {\color{red}0.1719}&{\color{red}0.1792}& {\color{red}0.1236}& \textbf{\color{red}0.0649}& {\color{red}0.1069} \\
\hline
 ROUGE-2 Precision & {\color{red}0.1999}& {\color{red}0.2641}& {\color{red}0.1670}& \textbf{\color{red}0.1974}& {\color{red}0.3578} \\
\hline
 ROUGE-2 F1 &{\color{red}0.1577} & {\color{red}0.1811}& {\color{red}0.1292}& \textbf{\color{red}0.0950}& {\color{red}0.1646} \\
\hline
 ROUGE-L Recall & {\color{ForestGreen}0.2973}& \textbf{\color{ForestGreen}0.2945}& {\color{ForestGreen}0.2535}& {\color{ForestGreen}0.1822}& {\color{ForestGreen}0.1928} \\
\hline
 ROUGE-L Precision & {\color{ForestGreen}0.4034}& \textbf{\color{ForestGreen}0.4512}& 
 {\color{ForestGreen} 0.3608}& {\color{ForestGreen}0.4068}& {\color{ForestGreen}0.5428} \\
\hline
 ROUGE-L F1 & {\color{ForestGreen}0.2980}& \textbf{\color{ForestGreen}0.3192}& {\color{ForestGreen}0.2747}& {\color{ForestGreen}0.2458}& {\color{ForestGreen}0.2846} \\
\hline
\end{tabular}
\end{center}
\caption{\label{tab:table-nameS} Summarization Metrics for selected Court}
\end{table}

\textit{\underline{Conclusion:}} We compute the ROUGE metrics for a sample of 25 cases from 5 courts with the highest case counts. We conclude that for the Summaarization Metrics:
\begin{itemize}
    \item \textbf{HKLDT} has the highest {\color{teal} ROUGE-1} Metrics [Unigram]
    \item \textbf{HKFC} has the lowest {\color{red} ROUGE-2} Metrics [Bigram]
    \item \textbf{HKCA} has the highest {\color{ForestGreen} ROUGE-L} Metrics [Longest Common Subsequence]
\end{itemize}

\subsection{Paragraph-level Judgment Classification Results}

Firstly, we summarize the results of our models in the following Table. We considered a sample of 1000 paragraphs with 4 classifiers for a 80-20 Train-Test Set Split, with the classes as described in Section 5.4. 

\begin{table}[H]
\begin{center}
\begin{tabular}{|m{5.7cm}|m{1.8cm}|m{1.8cm}|m{1.6cm}|m{1.8cm}|} 
\hline
\textbf{Model/Metric} & \textbf{Accuracy}&\textbf{Precision}&\textbf{Recall}&\textbf{F1 Score}\\  
\hline\hline
\textbf{Boltzmann Machine} & \textbf{\color{red}0.55} & \textbf{\color{red}0.30} & \textbf{\color{red}0.55} & \textbf{\color{red}0.39} \\
\hline
\textbf{Bernoulli Naive Bayes} & 0.73 & 0.67 & 0.73 & 0.69 \\
\hline
\textbf{Multinomial Naive Bayes} & 0.68 & 0.71 & 0.67 & 0.61 \\
\hline
\textbf{Complement Naive Bayes} & 0.72 & 0.73 & 0.72 & 0.69 \\
\hline
\textbf{Support Vector Machine} & 0.77 & 0.76 & 0.76 & 0.75 \\
\hline
\textbf{Logistic Regression} & 0.78 & 0.78 & 0.78 & 0.77 \\
\hline
\textbf{Multilayer Perceptron} & 0.74 & 0.74 & 0.74 & 0.73 \\
\hline
\textbf{Embedding+Linear} &  0.62  & 0.61 & 0.62 & 0.61 \\
\hline
\textbf{E(LSTM)+D(LSTM+Attn)} &  \textbf{\color{ForestGreen}0.98}  & \textbf{\color{ForestGreen}0.97} & \textbf{\color{ForestGreen}0.94} & \textbf{\color{ForestGreen}0.95} \\
\hline

\end{tabular}
\end{center}
\caption{\label{tab:table-nameP} Model Metrics for Classification (Test Set)}
\end{table}

Therefore, we see that Bernoulli Restricted Boltzmann Machines perform the worst while the proposed Encoder Decoder Style Architecture performs the best. 

\textit{\underline{Limitations:} The models however, don't generalize well for different samples of test-set. Therefore, we suggest using the BART-Large Model trained on the MultiGenre Natural Language Inference Dataset for better attention to linguistic substructures in long paragraphs. Our Encoder-Decoder architecture fails to perform for models with large paragraphs and the accuracy of the model reduces to 0.47 for paragraphs of size more than 100 tokens. As the pre-trained model allows upto 512 tokens, and our data doesn't have more than 473 tokens in a paragraph, it performs as a better generalization model.}

\textit{\underline{Conclusion:}} Therefore, we conclude that Zero-Shot Learning performs a more realistic examination of textual classification of legal data with a relatively fast inference time, due to its pretraining on multiple large datasets. The architecture of BART is described [48] as a Transformer based Encoder-Decoder style architectural language model with a Bidirectional Encoder and an Autoregressive Decoder. The model was pre-trained using corrupted text with an arbitrary noising functions making it learn to reconstruct the original text by denoising.

Some of the work done by BART prediction is shown in the following figure with highlights for the texts shown to cause the model to consider one type of class over the other done by human crosschecking. 

\begin{figure}[H]
\centerline{\includegraphics[]{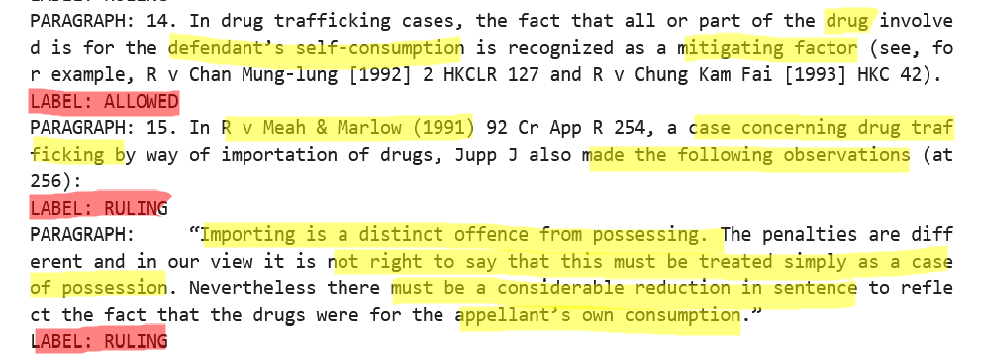}}
\caption{Why BART generalized better? (Part 1)}
\end{figure}

\begin{figure}[H]
\centerline{\includegraphics[]{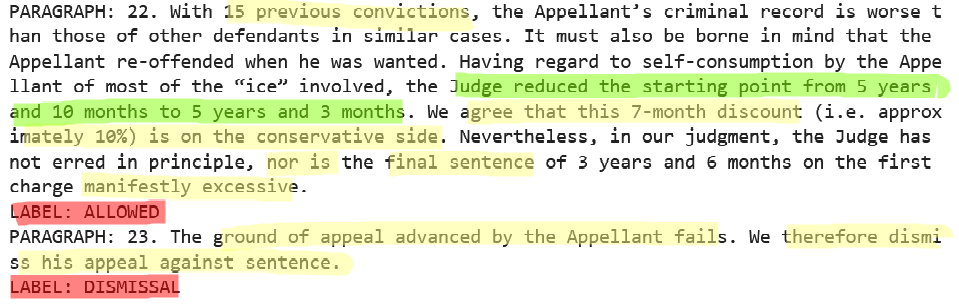}}
\caption{Why BART generalized better? (Part 2)}
\end{figure}

\subsection{Case Analysis 1: 7 HKCFAR 187}

For the following case, we show the keyword analysis results and the subsequent sentiment distribution paragraph wise with tagging.

\begin{table}[H]
\begin{center}
\begin{tabular}{|m{2cm}|m{12cm}|} 
 \hline
 \textbf{Method} & \textbf{Result}\\  
 \hline\hline

 {\sc TextRank} Summary & The crucial issue of principle in this appeal is whether the Secretary in determining the potential deportee’s torture claim in accordance with the policy is entitled to rely merely on UNHCR’s  unexplained rejection of refugee status for the person concerned, without undertaking any assessment of the claim. \\

 \hline
\end{tabular}
\end{center}
\caption{\label{tab:table-nameKA1i} Keyword Analysis and Summary: 7 HKCFAR 187}
\end{table}

\begin{table}[H]
\begin{center}
\begin{tabular}{|m{2cm}|m{12cm}|} 
 \hline
 \textbf{Method} & \textbf{Result}\\  
 \hline\hline

 {\sc TextRank} Keywords & ['ani',
 'unhcr',
 'refuge',
 'state',
 'secretari',
 'tortur',
 'reason',
 'mr',
 'legal',
 'deport',
 'deporte',
 'convent',
 'concern',
 'law',
 'nation high',
 'art',
 'person',
 'relev consider includ',
 'sri lanka'
 ] \\

 \hline

 RAKE & ['anxious consideration mr prabakar expressed',
 'november',
 'based',
 'acknowledging',
 'permissible course',
 'determining refugee status',
 'claimed protection',
 'september',
 'way without undertaking',
 'take unhcr',
 'secretary merely following unhcr',
 'omission',
 'suspected',
 'accordance',
 'refugee saying',
 'suffering whether physical'] \\

 \hline

 YAKE  & ['Secretary', 'UNHCR', 'refugee', 'Convention', 'torture', 'respondent',  'person',  'Director', 'Hong',
'Kong', 'claim',
'Sri', 'status',
'concerned',
'country',
'order',
'Lanka',
'Art',
'reasons',
'State']

\\

 \hline

 LDA & ['secretary',
 'refugee',
 'unhcr',
 'torture',
 'would',
 'respondent',
 'person',
 'convention',
 'claim',
 'status']

 \\

 \hline
\end{tabular}
\end{center}
\caption{\label{tab:table-nameKA1} Keyword Analysis and Summary: 7 HKCFAR 187 (Continued)}
\end{table}

\begin{figure}[H]
\centerline{\includegraphics[width=1.2\textwidth]{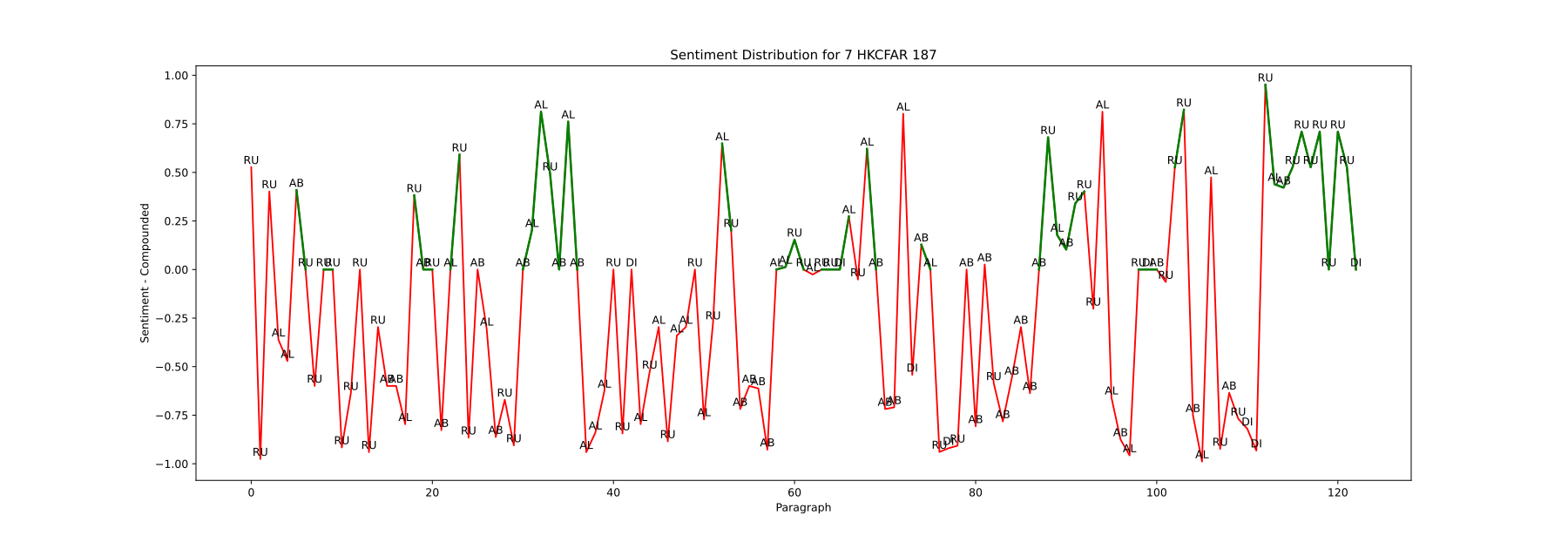}}
\caption{Sentiment Distribution with Tagging: 7 HKCFAR 187}
\end{figure}

\begin{figure}[H]
\centerline{\includegraphics[width=1.1\textwidth]{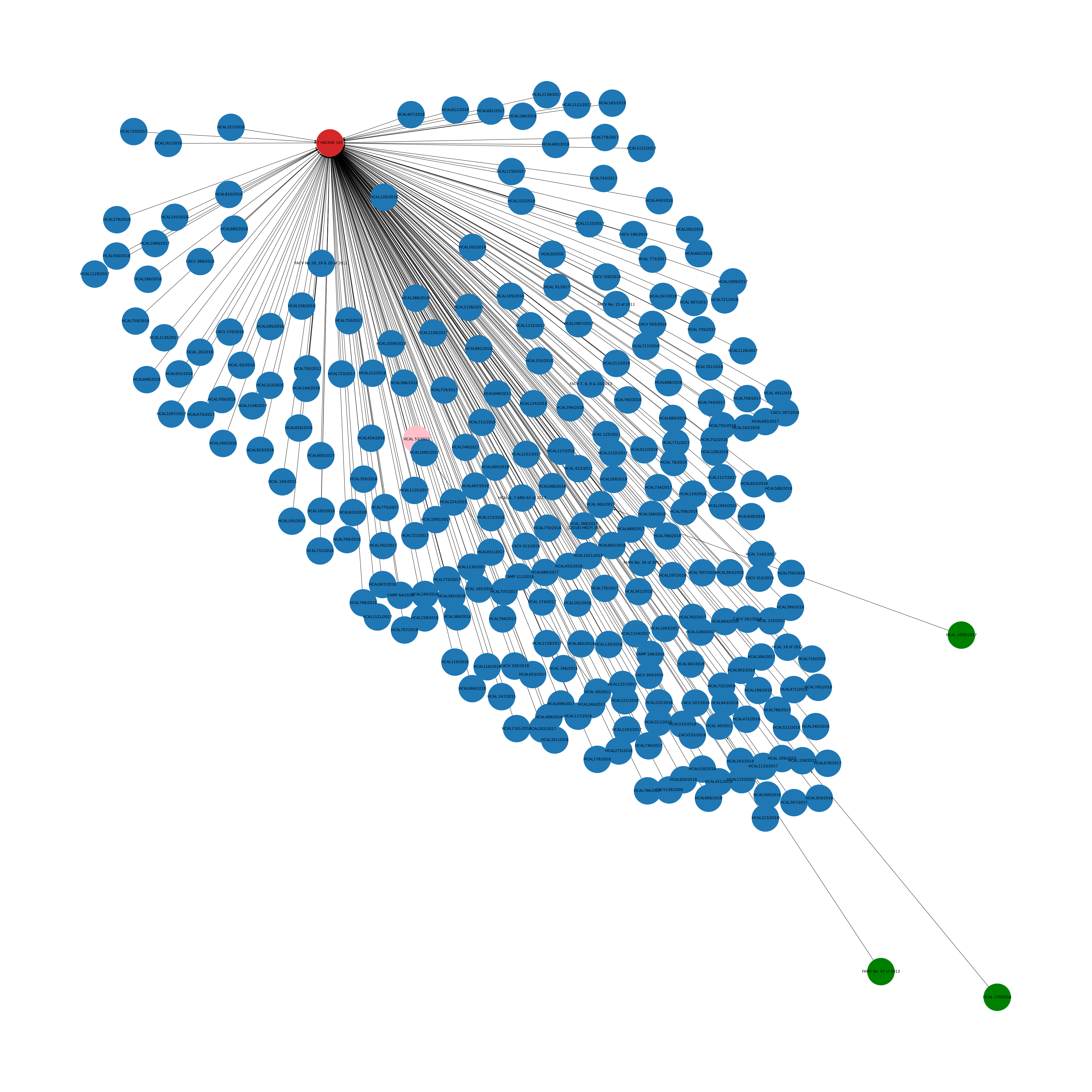}}
\caption{Citation Network Graph: 7 HKCFAR 187}
\end{figure}

\subsection{Case Analysis 2: 2 HKLRD 1121}

For the following case, we show the keyword analysis results and the subsequent sentiment distribution paragraph wise with tagging.

\begin{table}[H]
\begin{center}
\begin{tabular}{|m{2cm}|m{12cm}|} 
 \hline
 \textbf{Method} & \textbf{Result}\\  
 \hline\hline

 {\sc TextRank} Summary & The Judge emphasized the gravity of the offence of trafficking in dangerous drugs and pointed out that, according to the sentencing guidelines laid down by the Court of Appeal, the starting point for trafficking in up to 10 grammes of “ice” was 3 to 7 years' imprisonment. \\

 \hline
 
 {\sc TextRank} Keywords & ['sentenc',
 'judg',
 'drug',
 'ma',
 'charg',
 'appeal',
 'appel chow chun',
 'discount',
 'case',
 'offenc',
 'appropri',
 'defend',
 'consid',
 'consider',
 'onli',
 'polic',
 'fai hklrd',
 'traffick',
 'mitig'] \\

 \hline

 RAKE & ['judge considered',
 'defendant',
 'apparatus',
 'execution',
 'dangerous drugs',
 'guilty plea',
 'ning road',
 'court factual',
 'trafficking',
 'police officer',
 'appellant emphasized',
 'discount approximately',
 'inadequate',
 'months taking',
 'already sentenced',
 'take issue',
 'approached']
 \\

 \hline

 YAKE  & ['Appellant', 'Judge', 'Court', 'drug', 'years', 'month', 'ice', 'sentence', 'drugs', 'defendant', 'trafficking', 'Appeal', 'grammes', 'consumption', 'TMCC', 'point', 'starting', 'possession', 'discount', 'imprisonment']

\\

\hline

LDA & ['appellant',
 'months',
 'years',
 'drug',
 'judge',
 'ice',
 'sentence',
 'court',
 'defendant',
 'trafficking']

\\

 \hline
\end{tabular}
\end{center}
\caption{\label{tab:table-nameKA2} Keyword Analysis and Summary: 2 HKLRD 1121}
\end{table}

\begin{figure}[H]
\centerline{\includegraphics[width=1.2\textwidth]{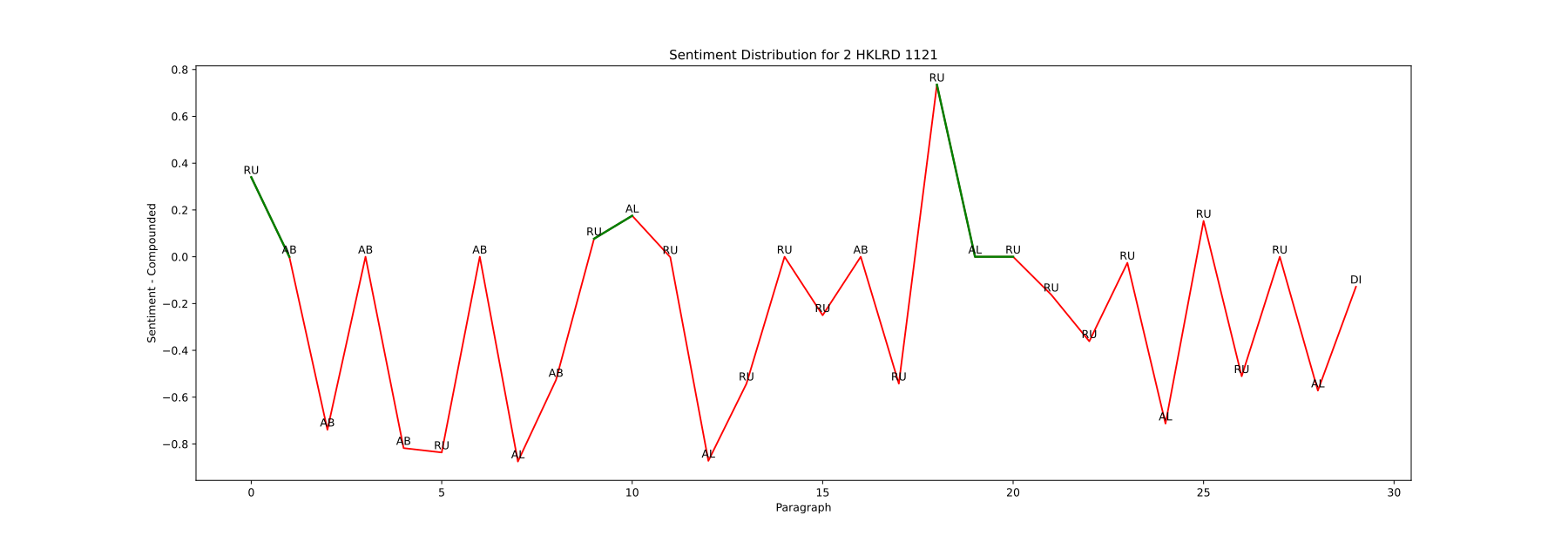}}
\caption{Sentiment Distribution with Tagging: 2 HKLRD 1121}
\end{figure}

\begin{figure}[H]
\centerline{\includegraphics[width=1.1\textwidth]{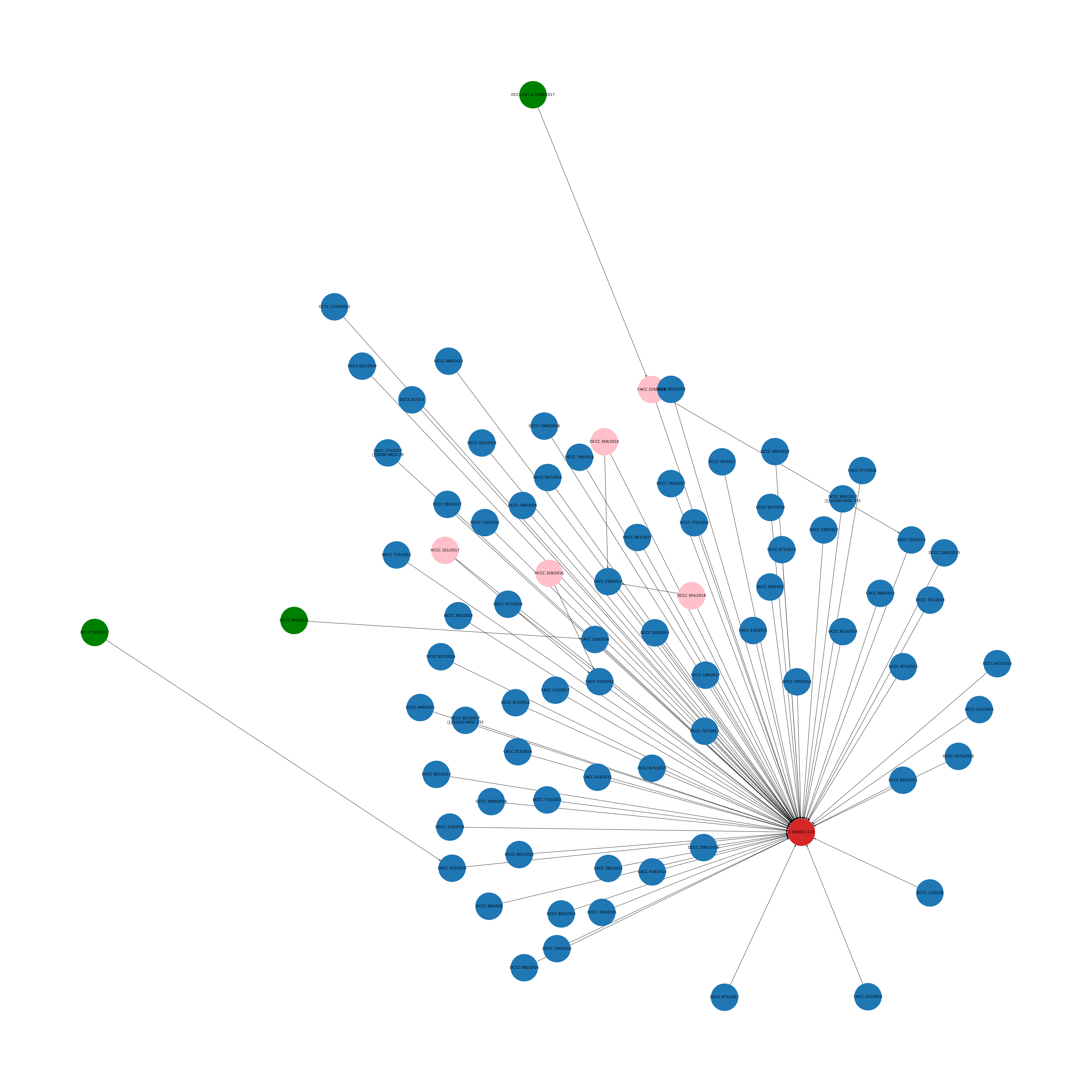}}
\caption{Citation Network Graph: 2 HKLRD 1121}
\end{figure}

\pagebreak

\section{Conclusion \& Future Works}

In this paper, we conducted extensive experiments and analysis for extraction of useful information from legal judgments from a computational linguistics viewpoint, with respect to judgments mete out by Hong Kong's Legal System. We implemented a 5-fold methodology: (1) Citation Network Graph Generation, (2) PageRank Algorithm, (3) Keyword Analysis and Summarization, (4) Sentiment Polarity, and (5) Paragrah Classification, for extraction of key insights from individual as well a group of judgments together, thus automating the extraction of useful insights with relatively fast inference times. We also coupled our experimental results by benchmarking our results using Large Language Models.

Our future work consists of finetuning two of the papers that we wrote as a complement to this project, and aim for submission to the upcoming International Conference of Legal Knowledge and Information Systems (JURIX 2023). We plan to upload the papers to arXiv in coming weeks, and we would appreciate feedback in relation to the papers. 

\begin{figure}[H]
\centering
\begin{subfigure}{0.5\textwidth}
  \centering
  \includegraphics[]{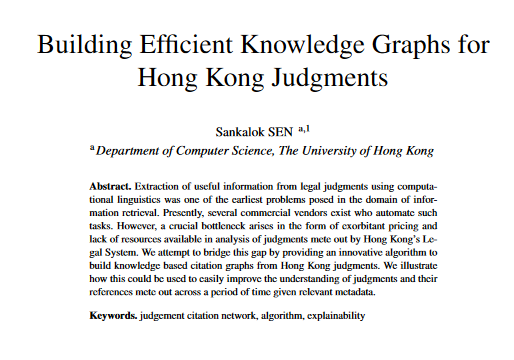}
  \caption{Citation Knowledge Graph}
  \label{fig:sub1}
\end{subfigure}%
\\
\begin{subfigure}{0.5\textwidth}
  \centering
  \includegraphics[]{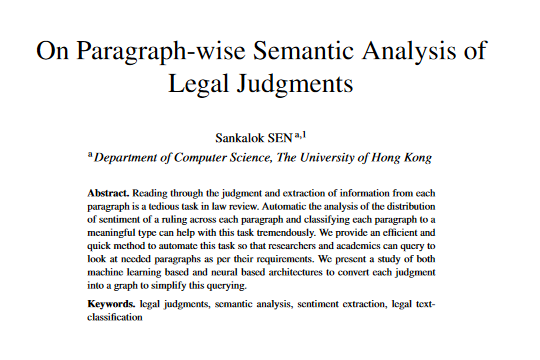}
  \caption{Paragraph-wise Semantic Analysis}
  \label{fig:sub2}
\end{subfigure}
\caption{Abstract of Papers for submission to JURIX, 2023}
\label{fig:test}
\end{figure}

\section{Acknowledgment}

I would like to thank my project advisor Dr. Lingpeng Kong for his continued support in this project to further my research interests in bridging the gap between Natural Language Processing and Social Science domains. I would also like to thank Dr. Zhiyong Wu of Shanghai AI Laboratory and Dr. Kevin Wu, Post Doctorate Fellow at The Department of Computer Science, HKU, for their helpful tips and suggestions in solving the multiple bottlenecks that this project aimed to tackle.

\pagebreak

\section{References}

\flushleft [1] S. H. C. Lo, K. K. Cheng, and W. H. Chui, The Hong Kong Legal System, Cambridge University Press, 2019.

\flushleft [2] J. Chan, and C. L. Lim, "From Colony to Special Administrative Region", in Law of the
Hong Kong Constitution, 2nd ed., Sweet Maxwell, 2015.

\flushleft [3] D. Locke, and G. Zuccon, "Case law retrieval: accomplishments, problems, methods and 
evaluations in the past 30 years", arXiv:2202.07209, 2022.

\flushleft [4] D. Bahdanau, K. Cho, and Y. Bengio, "Neural Machine Translation by Jointly Learning to
Align and Translate", In Proc. International Conference on Learning Representations (Oral 
Presentation), 2015.

\flushleft [5] A. Vaswani, N. Shazeer, N. Parmar, J. Uszkoreit, L. Jones, A. N. Gomez, L. Kaiser, and I. 
Polosukhin, "Attention Is All You Need", In Proc. 31st Conference on Neural Information 
Processing Systems, 2017.

\flushleft [6] D. Nguyen, N. A. Smith, C. P. Rose. 2011. Author Age Prediction from Text using Linear 
Regression. In Proceedings of the 5th ACL-HLT Workshop on Language Technology for 
Cultural Heritage, Social Sciences, and Humanities, pages 115–123. 

\flushleft [7] M. Sap, S. Swayamdipta, L. Vianna, X. Zhou, Y. Choi, N. A. Smith. 2021. Annotators 
with Attitudes: How Annotator Beliefs And Identities Bias Toxic Language Detection. arXiv 
preprint arXiv:2111.07997. 

\flushleft [8] S. Gururangan, D. Card, S. K. Dreier, E. K. Gade, L. Z. Wang, Z. Wang, L. Zettlemoyer, 
N. A. Smith. 2022. Whose Language Counts as High Quality? Measuring Language 
Ideologies in Text Data Selection. arXiv preprint arXiv:2201.10474. 

\flushleft [9] J. Gross, B. Acree, Y. Sim, N. A. Smith. 2013. Testing the Etch-a-Sketch Hypothesis: 
Measuring Ideological Signaling via Candidates’ Use of Key Phrases. In American Political 
Science Association Annual Meeting, Chicago 2013. 

\flushleft [10] E. G. Altmann, J. B. Pierrehumbert, A. E. Motter. 2011. Niche as a determinant of word 
fate in online groups. PLoS ONE 6(5).
19

\flushleft [11] A. Garimella, R. Mihalcea. 2016. Zooming in on Gender Differences in Social Media. In 
Proceedings of the Workshop on Computational Modeling of People’s Opinions, Personality, 
and Emotions in Social Media, pages 1–10.

\flushleft [12] L. Page, S. Brin, R. Motwani, and T. Winograd, "The PageRank Citation Ranking : Bringing Order to the Web", in WWW, 1999. 

\flushleft [13] R. Mihalcea and P. Tarau. 2004. TextRank: Bringing Order into Text. In Proceedings of the 2004 Conference on Empirical Methods in Natural Language Processing, pages 404–411, Barcelona, Spain. Association for Computational Linguistics.

\flushleft [14] R. Campos, V. Mangaravite, A. Pasquali, A. Jatowt, A. Jorge, C. Nunes, and A. Jatowt. 2020. YAKE! Keyword Extraction from Single Documents using Multiple Local Features. In Information Sciences Journal. Elsevier, Vol 509, pp 257-289.

\flushleft [15] S. Rose, D. Engel, N. Cramer, and W. Cowley. 2010. Automatic Keyword Extraction from Individual Documents. In Text Mining: Applications and Theory (pp.1 - 20)

\flushleft [16] C.J. Hutto and Eric Gilbert. 2014. VADER: A Parsimonious Rule-based Model for Sentiment Analysis of Social Media Text. In Proc. Eighth International Conference on Weblogs and Social Media (ICWSM).

\flushleft [17] S. Eshima, K. Imai, T. Sasaki. 2020. Keyword Assisted Topic Models. https://arxiv.org/abs/2004.05964.

\flushleft [18] D. M. Blei, A. Y. Ng, and M. I. Jordan. 2003. Latent Dirichlet Allocation. In The Journal of 
Machine Learning Research. 

\flushleft [19] J. Gamper. 2000. A parallel corpus of Italian/German legal texts. In Proc. LREC.

\flushleft [20] C. Grover, B. Hachey, and I. Hughson. 2004. The HOLJ corpus: supporting summarisation of legal texts. In Proc. COLING.

\flushleft [21] B. Fawei, A. Wyner, and J. Pan. 2016. Passing a USA national bar exam: a first corpus for experimentation. In Proc. LREC.

\flushleft [22] Y. Kano, M. Kim, M. Yoshioka, Y. Lu, J. Rabelo, N. Kiyota, R. Goebel, and K. Satoh. 2018. Coliee-2018: Evaluation of the competition on legal information extraction and entailment. In Proc. JSAI. 

\flushleft [23] P. H. L. de Araujo, T. E. de Campos, R. R. R. de Oliveira, M. Stauffer, S. Couto, and P.Bermejo. 2018.  Lener-br: A dataset for named entity recognition in brazilian legal text. In Proc. PROPOR.

\flushleft [24] C. Xiao, H. Zhong, Z. Guo, C. Tu, Z. Liu, M. Sun, Y. Feng, X. Han, Z. Hu, H. Wang, and J. Xu. 2018. CAIL2018: A Large-Scale Legal Dataset for Judgment Prediction.

\flushleft [25] L. Manor and J. J. Li. 2019. Plain English summarization of contracts. In Proc. Natural Legal Language Processing Workshop.

\flushleft [26] I. Chalkidis, I. Androutsopoulos, and N. Aletras. 2019. Neural Legal Judgment Prediction in English. In Proc. ACL. 

\flushleft [27] I. Chalkidis, M. Fergadiotis, P. Malakasiotis, and I. Androutsopoulos. 2019. Large-Scale Multi-Label Text Classification on EU Legislation. In Proc. ACL.

\flushleft [28] X. Duan, B. Wang, Z. Wang, W. Ma, Y. Cui, D. Wu, S. Wang, T. Liu, T. Huo, and Z. Hu. 2019. Cjrc: A reliable human-annotated benchmark dataset for chinese judicial reading comprehension. In proc. CCL.

\flushleft [29] B. Luo, Y. Feng, J. Xu, X. Zhang, and D. Zhao. 2017. Learning to predict charges for criminal cases with legal basis. In Proc. EMNLP. 

\flushleft [30] Y. Shen, J. Sun, X. Li, and L. Zhang, Y. Li, and X. Shen. 2018. Legal Article-Aware End-To-End Memory Network for Charge Prediction. In Proc. CSAE.

\flushleft [31] S. Long, C. Tu, Z. Liu, and M. Sun. 2018. Automatic Judgment Prediction via Legal Reading Comprehension. https://arxiv.org/abs/1809.06537.

\flushleft [32] J. Li, G. Zhang, H. Yan, L. Yu, and T. Meng. 2018. A Markov Logic Networks Based Method to Predict Judicial Decisions of Divorce Cases. In Proc. IEEE SmartCloud. 

\flushleft [33] H. Ye, X. Jiang, Z. Luo, and W. Chao. 2018. Interpretable Charge Predictions for Criminal Cases: Learning to Generate Court Views from Fact Descriptions. In Proc. NAACL-HLT.

\flushleft [34] Y. Wu, K. Kuang, Y. Zhang, X. Liu, C. Sun, J. Xiao1, Y. Zhuang, L. Si, and F. Wu. 2020. De-Biased Court’s View Generation with Causality. In Proc. EMNLP. 

\flushleft [35] C. Cardellino, M. Teruel, L. A. Alemany, and S. Villata. 2017. Legal NERC with ontologies, Wikipedia and curriculum learning. In Proc. EACL.

\flushleft [36] A. Elnaggar, R. Otto, and F. Matthes. 2018. Deep Learning for Named-Entity Linking with Transfer Learning for Legal Documents. In Proc. AICCC. 

\flushleft [37] E. Leitner, G. Rehm, and J. Moreno-Schneider. 2019. Fine-Grained Named Entity Recognition in Legal Documents. In Proc. SEMANTiCS.

\flushleft [38] N. Lagos, F. Segond, S. Castellani, and J. O'Neill. 2010. Event extraction for legal case building and reasoning. In Proc. IIP. 

\flushleft [39] M. Truyens and P. V. Eecke. 2014. Legal aspects of text mining. In Proc. LREC.

\flushleft [40] K. Raghav, P. K. Reddy, and V. B. Reddy. 2016. Analyzing the extraction of relevant legal judgments using paragraph-level and citation information. In Proc. ECAI.

\flushleft [41] V. Tran, M. L. Nguyen, and K. Satoh. 2020. Building legal case retrieval systems with lexical matching and summarization using a pretrained phrase scoring model. https://arxiv.org/abs/2009.14083.

\flushleft [42] C. Grover, B. Hachey, L. Hugson, C. Korycinski. 2003. Automatic summarisation of legal documents. In Proc. ICAIL.

\flushleft [43] R. Kumar V and K. Raghuveer. 2012. Legal Document Summarization using Latent Dirichlet Allocation. In Proc. IJCST. 

\flushleft [44] R. S. Wagh and D. Anand. 2020. A Novel Approach of Augmenting Training Data for Legal Text Segmentation by Leveraging Domain Knowledge. In Proc. Technologies and Applications 2020.

\flushleft [45] Hong Kong S.A.R., Hong Kong Legal Information Institute (HKLII), 
https://www.hklii.hk/eng/.

\flushleft [46] Hong Kong S.A.R. Legal Reference System, 
https://legalref.judiciary.hk/lrs/common/help/hlptopic.htm.

\flushleft [47] C.-Y. Lin. 2004. ROUGE: A Package for Automatic Evaluation of Summaries. In Text Summarization Branches Out, Association for Computational Linguistics.

\flushleft [48] Huggingface, https://huggingface.co/facebook/bart-large.

\end{document}